\documentclass[tinyml]{acmart}

\usepackage{caption}
\usepackage{subcaption}
\usepackage{siunitx}
\usepackage{multirow}
\catcode`\%=12\relax
\DeclareSIUnit[number-unit-product = ]\percent{\%}
\catcode`\%=14\relax

\AtBeginDocument{%
  \providecommand\BibTeX{{%
    \normalfont B\kern-0.5em{\scshape i\kern-0.25em b}\kern-0.8em\TeX}}}

\setcopyright{rightsretained}
\copyrightyear{2024}
\acmYear{2024}

\acmConference[tinyML Research Symposium'24]{tinyML Research Symposium}{April 2024}{San Francisco, CA}%

\begin{document}

\SetKwComment{Comment}{/* }{ */}

\title[Simulating Battery-Powered TinyML Systems Optimised using Reinforcement Learning]{Simulating Battery-Powered TinyML Systems Optimised using Reinforcement Learning in Image-Based Anomaly Detection}


\author{Jared M. Ping}
\email{jp140694@gmail.com}
\affiliation{%
  \institution{School of Electrical and Information Engineering, University of the Witwatersrand \& Digital Matter}
  \city{Johannesburg}
  \country{South Africa}
}

\author{Ken J. Nixon}
\email{ken.nixon@wits.ac.za}
\orcid{0000-0001-5391-8147}
\affiliation{%
  \institution{School of Electrical and Information Engineering, University of the Witwatersrand}
  \city{Johannesburg}
  \country{South Africa}
}


\begin{abstract}
Advances in Tiny Machine Learning (TinyML) have bolstered the creation of smart industry solutions, including smart agriculture, healthcare and smart cities. Whilst related research contributes to enabling TinyML solutions on constrained hardware, there is a need to amplify real-world applications by optimising energy consumption in battery-powered systems. The work presented extends and contributes to TinyML research by optimising battery-powered image-based anomaly detection Internet of Things (IoT) systems. Whilst previous work in this area has yielded the capabilities of on-device inferencing and training, there has yet to be an investigation into optimising the management of such capabilities using machine learning approaches, such as Reinforcement Learning (RL), to improve the deployment battery life of such systems. Using modelled simulations, the battery life effects of an RL algorithm are benchmarked against static and dynamic optimisation approaches, with the foundation laid for a hardware benchmark to follow. It is shown that using RL within a TinyML-enabled IoT system to optimise the system operations, including cloud anomaly processing and on-device training, yields an improved battery life of \SI{22.86}{\percent} and \SI{10.86}{\percent} compared to static and dynamic optimisation approaches respectively. The proposed solution can be deployed to resource-constrained hardware, given its low memory footprint of \SI{800}{\byte}, which could be further reduced. This further facilitates the real-world deployment of such systems, including key sectors such as smart agriculture.
\end{abstract}


\maketitle

\section{Introduction}

The rapid emergence of TinyML enables applications in several fields such as smart agriculture, healthcare, and smart cities \cite{Applications, AgriML}. Several TinyML applications within the smart agriculture sector leverage existing IoT hardware, specifically containing low-power Microcontroller Units (MCUs) constrained in resource and computational capability \cite{AgriTinyML}. These limited MCUs can now achieve advanced edge processing through optimised inferencing and on-device training \cite{MCUNetV3},  which yields security benefits by mitigating the need to transmit sensitive data over the air.

These advancements unlock value in existing low-cost hardware but present a new challenge as many such systems utilise a finite battery power supply that would be depleted faster due to the increased computation \cite{Applications}. An existing battery-powered TinyML IoT data-logging system, periodically uploading results, intended to last many years in the field, now equipped with the additional capabilities to perform inferencing and re-training on new data collected, would last a fraction of its original deployment battery life. Hence, there is a requirement for an intelligent optimisation scheme to orchestrate the balance between function and limited power supply within such solutions to improve the viability of real-world deployments.

This paper presents and evaluates the simulated impact of utilising RL as a means of autonomous optimisation compared to traditional static and dynamic optimisation schemes. Static optimisation is performed at compile or load time. This infers that the optimisation is performed once, and the results are applied for the entirety of the system deployment. Dynamic optimisation is better suited to solutions where behaviour adoption is required or environmental conditions differ from expectations. Autonomous optimisation is a mechanism of dynamic optimisation whereby the optimal action is self-learned rather than explicitly dictated by the algorithm.

The autonomous approach implements a decayed epsilon-greedy Q-learning, off-policy, RL algorithm within an image-based anomaly detection system, which enables the system to determine the optimal action by tracking the remaining battery capacity, number of classified anomaly images, and anomaly status per sample \cite{Bellman, EpsilonGreedy}. Q-learning is well-suited due to the discrete nature of the limited state information being tracked within such a system \cite{RLSurvey}. This translates to a small memory footprint, allowing it to scale to many resource-constrained MCUs typically found within TinyML-capable IoT solutions. The algorithm can be pre-trained and supports ongoing learning to enable adaptation to changing or new environments. A generic reward function allows the algorithm to be hardware agnostic, provided hardware meets the minimum resource and functional requirements \cite{Reward}.

The hardware and environment are simulated for a battery-powered TinyML image-based anomaly detection IoT system to validate this solution. Hardware simulation involves selecting commonly available components that may feature in such a system, which will be validated through a forthcoming physical experiment. The system supply is constrained to an ideal finite battery supply, with the battery being modelled to have a predefined capacity and linear discharge characteristics. The environment is simulated through control of varying anomaly occurrence ratios, ranging from \SI{5}{\percent} to \SI{40}{\percent}. Re-training behaviour is modelled and varies according to the number of input data samples supplied. Variations in environmental conditions are constrained and reliably reproduced to facilitate accurate comparison between optimisation methods.

This research provides the following valuable contributions to the field of TinyML:

\begin{itemize}
    \item Modelling and implementing static, dynamic, and autonomous optimisation algorithms for a TinyML image-based anomaly detection IoT system that supports on-device inferencing and training as well as cloud-based inferencing.
    \item The battery life impact of the autonomous optimisation algorithm has been measured against static and dynamic optimisation algorithms and verified through detailed simulation, yielding an improvement of \SI{22.86}{\percent} and \SI{10.86}{\percent} respectively.
\end{itemize}

\section{Related Work}

\begin{figure}[htbp]
    \centering
    \includegraphics[width=\linewidth]{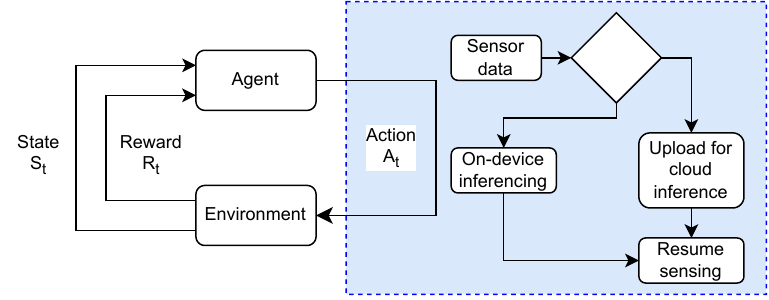}
    \caption{Reinforcement learning state machine with the inclusion of optimisation actions to balance on-board power consumption versus cloud processing latency in prior research.}
    \label{fig:prior-state}
    \Description{Flow diagram depicting how reinforcement learning was utilised to balance on-device inferencing versus uploading to the cloud for fog inferencing.}
\end{figure}

Research conducted by Liu et al. in 2019 \cite{QOffload} proposes a Q-greedy reinforcement algorithm to optimise the power versus latency trade-off of a system with on-device processing and cloud offloading capabilities. The proposed algorithm utilises a cellular medium to perform network uploads. The scope of this work is illustrated in Figure~\ref{fig:prior-state}. The result of this previous work yields a small improvement in optimising the average power consumption of the edge devices through an RL optimisation scheme whilst improving offloading task latency in tandem.

This provides a new research opportunity to extend the proposed algorithm to include the capability of on-device training within the action set. On-device training consumes much energy and requires optimal control to ensure it is performed sparingly to prevent a drastic reduction in the deployment battery life.

Research conducted by Ren et al. in 2019 complements the aforementioned work by considering that offloading computation incurs transmission delays and requires available cloud capacity for such computation to be processed off-device\cite{RenFederated}. This demonstrates the complexity involved in offloading and the significance of reducing this off-device compute load, where possible, such as through on-device training, although this has other benefits, such as reducing energy usage on frequent offloading uploads.

Research conducted by Basaklar et al. in 2022 highlights the applicability of utilising RL in the scope of TinyML action optimisation by considering the balance in responding to shifting environmental conditions whilst achieving functional objectives on systems with limited battery power and computational capacity\cite{Tinyman}.

\section{Optimising with Reinforcement Learning}\label{sec:models}

Core motivations and objectives are defined in this section. Static, dynamic and autonomous optimisations are explored and algorithms are proposed for comparison. A brief overview of the autonomous algorithm selection criteria and trade-offs is discussed before detailing the implementation of the autonomous RL optimisation scheme.

\subsection{Problem Definition}

A large portion of TinyML research has contributed to enabling edge functionality on resource-constrained hardware through optimisation, including inferencing and training \cite{MCUNetV1, MCUNetV2, MCUNetV3, TinyMLSurvey, TinyMLSurvey2}. Many proposals leverage machine learning optimisation techniques, specifically RL, to achieve this goal \cite{RLSurvey, QOffload, RLDataAugment}. There is demonstrated value in utilising RL to optimise the orchestration of constrained hardware that utilises on-device inferencing and training.

The optimisation problem shifts from the prior work illustrated in Figure~\ref{fig:prior-state}, of balancing latency and power consumption of on-device versus fog inferencing. The current problem is balancing environmental image-based anomaly uploads versus performing on-device training to remove a continued upload requirement and conserve remaining system energy. The choice of upload medium relies on the data bandwidth and latency requirements. This solution uses NB-IoT as a wireless medium, as processing latency is not a priority, and data bandwidth is sufficient for small image uploads.

\begin{figure}[H]
    \centering
    \includegraphics[width=\linewidth]{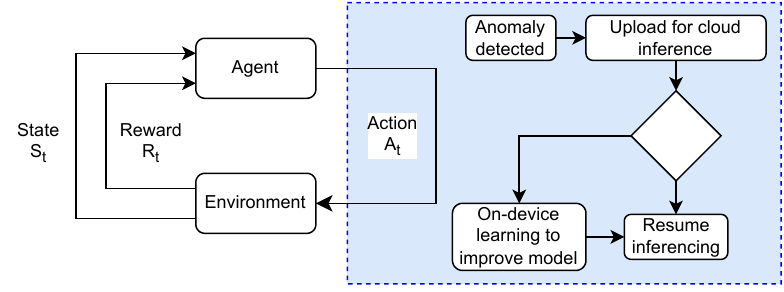}
    \caption{Reinforcement learning state machine with the inclusion of actions to be optimised to extend system deployment battery life in the current scope of research.}
    \label{fig:current-state}
    \Description{Flow diagram depicting how reinforcement learning will be utilised to balance on-device training versus uploading to the cloud for fog inferencing to enable on-device training.}
\end{figure}

On-device models need to be lightweight to fit within the limited MCU resources, and hence, a cloud server exists with a larger model to classify the locally inferred anomalies. Cloud-classified anomalies are returned to the edge system to provide direction on whether the anomaly is valid or classified into an identifiable class. Achieving optimised on-device training enables real-world application of such functionality. This new scope of work is illustrated in Figure~\ref{fig:current-state}.

Research has been performed to select a candidate RL algorithm to optimise such a system and simulate the battery life impacts compared to non-machine learning approaches. Given a minimum viable hardware implementation of such a system, it is clear that the optimisation should focus on optimising the balance of, and hence total energy consumed by, both anomaly upload and on-device training actions since it can be generally stated that:

\begin{equation} 
E_{upload} \gg E_{image\_capture}, E_{image\_infer (on-device)}
\end{equation}
\begin{equation} 
E_{train} \gg E_{image\_capture}, E_{image\_infer (on-device)}
\end{equation}

\subsection{Static and Dynamic Approaches}\label{sec:stat-dyn}

This approach is typically a good candidate for IoT systems that are functionally simple, and it relies on environmental conditions being known or assumed before deployment \cite{EdgeOptimisation}. Therefore, the optimisation yields are thwarted when environmental variances occur. These changes could be accounted for by delivering system updates over the air. However, this requires ongoing monitoring of the environmental conditions and is generally not perceived as scalable or maintainable. A candidate static algorithm is proposed in Algorithm~\ref{alg:static}, with initial thresholds defined based on the simulated environment behaviour in Section~\ref{sec:sim-env}.

\begin{algorithm}
\caption{Static Optimisation Algorithm}\label{alg:static}
$N \gets 0$ \Comment*[r]{Number of classified anomalies}
$T \gets 35$ \Comment*[r]{Required training threshold}
$V \gets 0.85$ \Comment*[r]{Training validation threshold}

\If{N >= T}
{
    $R \gets TrainAndValidate()$ \Comment*[r]{Validation acc.}

    \If{R >= V}
    {
        $N \gets 0$\;
    }
}
\end{algorithm}

This approach allows the system to maximise the intended optimisation yields in response to a range of environmental changes, though some may not have been anticipated at design time, by analysing the current state of the system and the anticipated future states \cite{EdgeOptimisation}. This optimisation does have its drawbacks, as the algorithm may lag in responding to certain environmental conditions. Furthermore, the complexity of designing and implementing such an all-encompassing algorithm can be significantly more complex than a static approach. A candidate dynamic algorithm is proposed in Algorithm~\ref{alg:dynamic}, with initial thresholds defined based on the simulated environment behaviour in Section~\ref{sec:sim-env}.

\begin{algorithm}
\caption{Dynamic Optimisation Algorithm}\label{alg:dynamic}
$N \gets 0$ \Comment*[r]{Number of classified anomalies}
$T \gets 10$ \Comment*[r]{Required training threshold}
$V \gets 0.85$ \Comment*[r]{Training validation threshold}
$S \gets 0$ \Comment*[r]{Successful training iterations}
$Z \gets 5$ \Comment*[r]{Data set reduction threshold}

\If{N >= T}
{
    $R \gets TrainAndValidate()$ \Comment*[r]{Validation acc.}

    \eIf{R >= V}
    {
        $N \gets 0$\;
        $S \gets S + 1$\;

        \If{S >= Z}
        {
            $T \gets T - 1$\;
        }
    }
    {
        $S \gets 0$\;
        $T \gets T + 1$\;
    }
}
\end{algorithm}

\subsection{Autonomous Approaches}

This approach utilises the dynamic state space of the environment in conjunction with the available actions to learn the optimal approach rather than having the strategy explicitly defined. This can reduce the design complexity compared to a dynamic optimisation algorithm and better adjust to unanticipated events that the dynamic algorithm design has not considered. This optimisation approach can incur large computation and energy costs as the agent needs to experience the environmental changes to learn the optimal solution \cite{RLSutton}.

Considering the target system in question, the state includes the current remaining battery life, the latest sample anomaly status, and the data available to re-train the device model. The actions include continuing anomaly uploads to the cloud or performing on-device re-training. Three candidate algorithms are identified in order of candidacy:

\textbf{Q-learning:} An off-policy RL algorithm that learns a value function, estimating the expected reward per action in a particular state \cite{Qlearning}. The value function is learned through trial and error, through rewarding the agent for taking actions that maximise the overall battery life of the system.

\textbf{Policy gradient:} An RL algorithm that learns a policy which maps states to relevant actions \cite{SuttonPolicyGradient}. The algorithm optimises the reward function, which can incur a larger computational overhead than Q-learning. However, this can yield a better solution to a frequently changing environment.

\textbf{Actor-critic:} A hybrid RL algorithm that combines both the Q-learning and policy gradient approaches to yield the advantages of both \cite{ActorCritic}. The critic uses Q-learning to learn the value function, while the actor uses policy gradient to learn a well-suited policy. The trade-off is that the implementation thereof can be significantly more complex.

\subsection{Proposed Autonomous Scheme}

Due to RL algorithms being computationally intensive by nature, the available computational resources should be considered when selecting a candidate algorithm \cite{Compute}. Classical Q-learning is chosen as the primary evaluation candidate as it yields optimal results within a small dataset and is well suited to a small discrete environment \cite{Qlearning, QSample}. This solution is well suited within a single-agent environment, and the moderate number of state-action pairs translates to a small memory footprint that many resource-constrained MCUs can support \cite{Qlearning}. A decayed epsilon-greedy policy is applied to this algorithm to maximise initial exploration during training before eventually exploiting what has been learnt \cite{RLSutton}. Training versus ongoing algorithm usage bears implementation resemblance, where the former utilises higher exploration rates to populate the Q-table with informed state-action pairs. The evaluation, described in Section~\ref{sec:bench}, has yielded sufficient improvements that the remaining candidates have not been evaluated. This implementation is described in Algorithm~\ref{alg:autonomous}.

\begin{algorithm}
\caption{Initial Q-Table Training Algorithm}\label{alg:autonomous}
$\alpha \gets 0.2$ \Comment*[r]{Learning rate}
$\gamma \gets 0.95$ \Comment*[r]{Discount factor}
$\epsilon \gets 1$ \Comment*[r]{Initial exploration rate}

 Initialise Q-table $Q(s, a)$ state-action pairs $(s, a)$ to zero\;
 \For{each episode}
 {
    Initialise state $s$\;
    \While{$s$ is not terminal}
    {
        $\alpha \gets \alpha \times 0.995$\;
        Pick action $a$ for state $s$ via decayed $\epsilon$-greedy policy:\\
        \Indp$\epsilon \gets \epsilon \times 0.99$\;
        $rnd \gets$ random number between 0 and 1\;
        \Indm\eIf{$rnd < \epsilon$}
        {
            $a \gets$ choose random action from possible actions;
        }
        {
            $a \gets argmax$ over $a$ from $Q(s, a)$\;
        }
        Do action $a$, observe reward $r$ and next state $s'$\;
        Update Q-value for state-action pair $(s, a)$:\\
        \Indp$Q(s, a) \leftarrow Q(s, a) + \alpha \left( r + \gamma \max_{a'} Q(s', a') - Q(s, a) \right)$\;
        \Indm$s \leftarrow s'$\;
    }
 }
\end{algorithm}

\section{Evaluation} \label{sec:eval}

The proposed autonomous optimisation scheme is benchmarked against both proposed static and dynamic algorithms for each chosen anomaly occurrence ratio is evaluated in this section. This includes simulated system definition, modelled consumption properties, and emulated re-training behaviour. Thereafter, the cumulative energy consumption per action is compared, algorithm behaviour is explored, and the overall battery life improvement of the system is compared.

\subsection{Simulated System and Environment}\label{sec:sim-env}

\begin{figure}[tbp]
    \centering
    \includegraphics[width=0.75\linewidth]{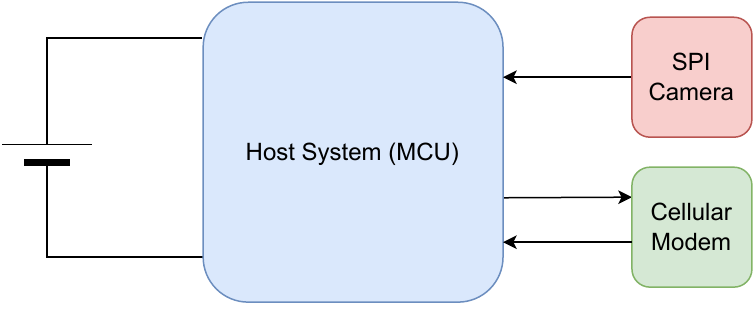}
    \caption{Block diagram of simplified image-based anomaly detection TinyML IoT system to be simulated.}
    \label{fig:system}
    \Description{Functional block diagram for the high-level system includes the MCU, power supply, SPI camera, and cellular modem.}
\end{figure}

The core system components are illustrated in Figure~\ref{fig:system}, required to implement the states and actions detailed in Table~\ref{tab:state-hardware}. These represent all significant contributors to energy utilisation during system operation.

The simulated system is a theoretical hardware implementation of the system illustrated in Figure~\ref{fig:system}. The functional components and corresponding simulated hardware are described in Table~\ref{tab:state-hardware}. This includes the modelled energy consumption per state and action extracted from the relevant datasheets, including reference to prior benchmark results for timing results of specific actions \cite{MCUNetV3}. The model utilised for benchmarking the training and inference time is a visual wake words model requiring 478~KB and 190~KB of MCU flash and RAM, respectively. There is demonstrated value in providing benchmarks that utilise existing hardware \cite{PeopleCount}. The selected simulated components offer a balance of cost versus power efficiency for the required system, criteria that would apply to any real-world design.

\begin{table}[tbp]
  \caption{States and actions supported by the system and simulated system hardware including energy consumption}
  \label{tab:state-hardware}
  \begin{tabular}{llll}
    \toprule
    State/Action & \multicolumn{1}{p{1.2cm}}{\centering Responsible\\Component} & \multicolumn{1}{p{1.2cm}}{\centering Simulated\\Hardware} & $E_{avg}$ \\
    \midrule
    System Active & MCU & STM32F746xx & \SI{400}{\milli\watt\hour} \\ 
    System Sleep & MCU & STM32F746xx & \SI{50}{\micro\watt\hour} \\ 
    Infer Result & MCU & STM32F746xx & \SI{17}{\micro\watt\hour} \\
    Re-training & MCU & STM32F746xx & \SI{556}{\micro\watt\hour/img} \\
    Image Capture & SPI camera & OV2640 & \SI{180}{\micro\watt\hour} \\
    Anomaly Upload & Cell modem & SIM7000E & \SI{3}{\milli\watt\hour} \\
  \bottomrule
\end{tabular}
\end{table}

The power supply of such a system can vary substantially depending on the cost, lifespan, and housing constraints of the project. Furthermore, there is the consideration of a single-use battery versus a rechargeable solution, whereby the latter may be complemented by a renewable power source such as solar. For simplicity, the simulated system emulates a single-use ideal battery solution offering \SI{5}{\volt} over a battery capacity of \SI{3.5}{\ampere\hour}, yielding \SI{17.5}{\watt\hour} of deliverable energy. The ideal source was chosen to remove the complexities of modelling the non-linear discharge characteristics of different battery chemistries.

The SIM7000E cellular modem supports multiple radio access technologies. However, the energy consumption is modelled specifically for NB-IoT, a common industry technology utilised in IoT deployments \cite{NBIoT}. While obtaining a Global Navigation Satellite System (GNSS) fix in a deployed system may be required, this requirement has been excluded to simplify simulations. It should be noted that this requirement can be easily included at a later time due to the SIM7000E module containing an integrated GNSS module \cite{SIMCOM}.

Several varying factors may contribute to variance in model accuracy when performing re-training. However, considering that an existing model has been selected and successfully trained with acceptable accuracy for the initial deployment, the largest validation accuracy impact factor when performing re-training is the size and quality of the data set being utilised, respectively.

Assuming the quality of sourced data is sufficiently consistent, the re-training behaviour can be modelled to vary according to the input data set size. This simulated modelling accounts for anticipated variance in smaller data sets, which slowly diminishes as the data set size grows, given a more reliable re-training outcome is more likely. Whilst the simulated modelling behaviour may still differ from the behaviour of real-world model training, this modelling allows for such a variable to be constrained for a fair benchmark per optimisation scheme. This modelling is described in Algorithm~\ref{alg:re-train}.

\begin{algorithm}
\caption{Re-training validation and energy modelling algorithm}\label{alg:re-train}
\KwData{$N\_samples, E\_train\_image$}
\KwResult{$V\_accuracy, E\_consumed$}

$V\_accuracy \gets 0$ \Comment*[r]{Validation accuracy}
$E\_consumed\gets 0$ \Comment*[r]{Re-train energy}

$T\_offset \gets \frac{2 \times rand() - 1}{10} \times 0.95 ^ {N\_samples};$

$T\_result \gets \frac{1}{(1 + e^{-0.6 \times \log(N\_samples)})} - \frac{0.4}{N\_samples} - T\_offset ;$

\uIf{V\_accuracy > 1}
{
    $V\_accuracy \gets 1;$
}
\ElseIf{V\_accuracy < 0}
{
    $V\_accuracy \gets 0;$
}

$E\_consumed \gets E\_train\_image \times N\_samples;$
\end{algorithm}

\begin{figure}[htbp]
    \centering
    \includegraphics[width=\linewidth]{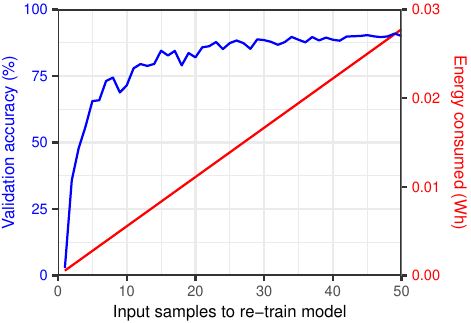}
    \caption{Simulated modelling of anticipated validation accuracy and energy required when performing re-training.}
    \label{fig:re-train}
    \Description{Graph showing the decreasing variance in anticipated model validation accuracy when a larger set of input samples are provided for the training.}
\end{figure}

The algorithm considers the estimated energy required to perform the re-training based on the size of the data set being utilised. This is computed by referencing the average MCU power consumption at the maximum configured frequency whilst considering the average time to re-train per image. The resulting simulated model is illustrated in Figure~\ref{fig:re-train}.

\subsection{Benchmarks and Results} \label{sec:bench}

Several simulated benchmarks have been performed using each optimisation algorithm described in Section~\ref{sec:models}. The variance between each benchmark test case is introduced through an adjusted anomaly detection ratio, namely the probability that a captured image yields an anomaly classification after inputting the trained onboard neural network. The variations of anomaly detection ratio are set in order of anticipated likeliness, based on the expectation of slow-changing environment conditions, namely \SI{5}{\percent}, \SI{10}{\percent}, \SI{20}{\percent} and \SI{40}{\percent}. The anomalies are artificially inserted probabilistically to ensure the target ratios are achieved.

\begin{figure}[tbp]
    \centering
    \includegraphics[width=\linewidth]{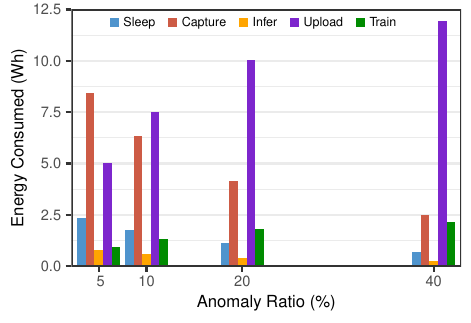}
    \caption{Simulated energy consumption per state and action within static optimisation benchmark.}
    \label{fig:static-operations}
    \Description{Bar graph illustrating the factor of energy consumed per action in the simulated system whilst using static optimisation, namely sleep, image capture, inference, upload, and train.}
\end{figure}

The benchmarks are designed to monitor all states and actions described in Table~\ref{tab:state-hardware}. This ensures that the system battery life is fairly modelled as parasitic energy consumers, such as the sleep state, account for a significant portion of consumed battery life over the deployment of a system. The impact of sleep state consumption is generally larger when the deployment lifetime is longer, as illustrated in Figure~\ref{fig:static-operations}.

\begin{figure*}[t]
\centering
\begin{subfigure}{1\columnwidth}
\includegraphics[width=\columnwidth]{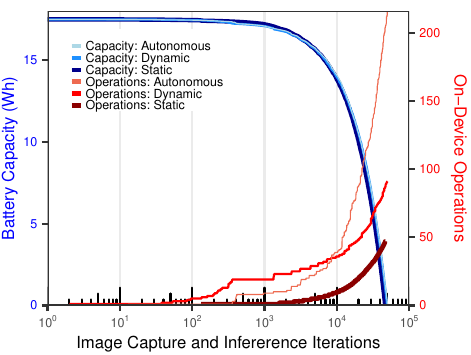}
\caption{Anomaly ratio of \SI{5}{\percent}.}
\label{fig:train-static}
\end{subfigure}
\hfill
\begin{subfigure}{1\columnwidth}
\includegraphics[width=\columnwidth]{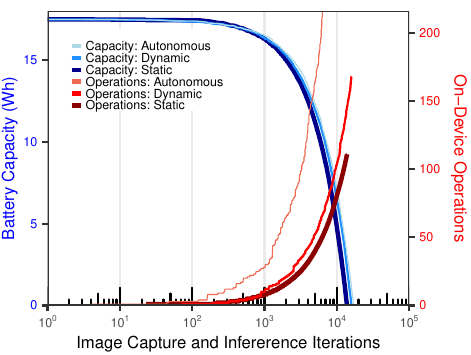}
\caption{Anomaly ratio of \SI{40}{\percent}.}
\label{fig:train-dynamic}
\end{subfigure}
\caption{Simulated energy utilisation through varying optimisation schemes to improve re-training attempt frequency.}
\label{fig:train-compare}
\Description{Two graphs showing the variances in training behaviour between the static, dynamic, and autonomous optimisation schemes at different anomaly ratios.}
\end{figure*}

A delimiter of 50 anomaly classifications is introduced to standardise the behaviour of anomaly classification after a successful re-training operation has taken place. This dictates the number of future anomalies that can be successfully identified onboard before a subsequent anomaly is treated as unknown and will require the results to be uploaded for server classification. Whilst this realistically varies in the field, ensuring the benchmarks offer a fair comparison with what is effectively treated as a shared seed for anomaly identification validity is crucial.

Each benchmark ensures system operation under the same conditions whereby the system is configured to wake hourly to capture an image, thereafter feeding the captured image into the onboard neural network for anomaly detection. If an anomaly is detected, an upload is performed, yielding an emulated classification from the server. The optimisation algorithm then determines whether a sufficient number of anomaly classifications has been obtained to attempt re-training the on-device neural network before returning to sleep. This process repeats until the available energy of the battery has been exhausted.

Each loop is defined as a \textit{sampling iteration}. Given the fixed frequency of sampling iterations, the total achievable iterations are translated into a simulated battery life per deployment on a single full charge of the emulated battery power supply. The obtained results from the respective benchmarks are tabulated in Table~\ref{tab:sim-benchmarks}.

\begin{table}[b]
  \caption{Simulated benchmark results comparing anomaly ratios and their respective impact per optimisation algorithm}
  \label{tab:sim-benchmarks}
  \begin{tabular}{cccc}
    \toprule
    \multicolumn{1}{p{1.2cm}}{\centering Anomaly\\Ratio\\(\si{\percent})} & \multicolumn{1}{p{1.8cm}}{\centering Static\\Battery Life\\(hr)} & \multicolumn{1}{p{1.8cm}}{\centering Dynamic\\Battery Life\\(hr)} & \multicolumn{1}{p{1.8cm}}{\centering Autonomous\\Battery Life\\(hr)} \\
    \midrule
     5  & 45~956 & 49~728 & 53~292 \\
     10 & 34~741 & 38~162 & 41~988 \\
     20 & 22~909 & 26~304 & 29~809 \\
     40 & 13~781 & 15~909 & 19~137 \\
     \midrule
     Avg. & 29~347 & 32~526 & 36~057 \\
  \bottomrule
\end{tabular}
\end{table}

Overall, the autonomous optimisation yields a deployment battery life improvement of \SI{22.86}{\percent} and \SI{10.86}{\percent} respectively when compared to the static and dynamic optimisation approaches respectively. The simulated system collecting samples at an hourly rate translates to a single deployment battery life of 4.11~years compared to the 3.35~years and 3.71~years of the static and dynamic systems, respectively. Although the autonomous optimisation scheme supports ongoing learning, it is found that further improvements in battery life during change of anomaly ratios are mitigated by the additional training. A larger year differential in the total deployment time will be seen if the frequency of the sampling interval is reduced from the simulated \SI{1}{\hour} period. The variance in device actions is illustrated in Figure~\ref{fig:train-compare}, demonstrating how the actions are optimised by each algorithm when the system experiences different anomaly occurrence ratios. This behaviour is subject to fine-tuning of the hyperparameters utilised by Algorithm~\ref{alg:autonomous}, which requires iterative evaluation to achieve maximised Q-table population.

\balance
\subsection{Deployability}

The simulated modelling and evaluation of such a solution highlights the portability of such a solution as it requires no specialised hardware and features a minuscule memory footprint of \SI{800}{\byte}. Pilot deployments will yield maximised value by allowing sufficient data collection of the initial environmental conditions. The initial training can then occur on cloud infrastructure, with the resulting Q-table deployed to edge deployments thereafter. The autonomous algorithm supports ongoing learning, allowing dynamic environmental conditions to be further accounted for and the learnings shared with deployments in similar environments.

The choice of simulated hardware for the benchmarks has been made with the intent to acquire the specified components and validate the simulated improvements through a physical experiment. Such a system is well-suited for use cases like those within smart agriculture, where varying visual conditions across deployments are probable.

Whilst this work does not include data privacy considerations, it should be noted that uploading anomaly data to the cloud should be done securely for any real-world deployment. The method in which this is achieved may have a battery life impact, such as encryption and decryption of anomaly data and server classification results.

\section{Conclusion}

The functional capabilities of TinyML devices continue to grow and enhance the value of several industries and sectors, such as smart agriculture. However, the unlocked enhancements can incur an increased power consumption profile that should be carefully managed for battery-powered systems. In working towards this goal, this paper presents the battery life benefits of utilising reinforcement learning, specifically Q-learning, to coordinate the operational management of image-based anomaly detection IoT systems capable of supporting on-board training to maximise the battery life per deployment. The solution is scalable due to its small memory footprint, requiring \SI{800}{\byte}. The proposed solution achieves \SI{22.86}{\percent} and \SI{10.86}{\percent} more battery life compared to the benchmarked static and dynamic algorithms respectively. Future work includes the scope of a physical experiment to validate the simulated improvements on actual components chosen for the simulation.

\bibliographystyle{ACM-Reference-Format}
\bibliography{references}




\end{document}